\documentclass[times,twocolumn,final]{elsarticle}
\usepackage{tikz}
\usepackage{cag}
\usepackage{framed,multirow}

\usepackage{amssymb}
\usepackage{latexsym}

\usepackage{url}
\usepackage{xcolor}
\definecolor{newcolor}{rgb}{.8,.349,.1}

\usepackage[hidelinks,colorlinks=false]{hyperref}
\usepackage{algpseudocode}
\usepackage{algorithm}
\usepackage{amsmath}
\usepackage{balance}
\usepackage{flushend}
\usepackage[switch,pagewise]{lineno} 
\newcommand{\etal}{\textit{et al.}}
\journal{Computers \& Graphics}

\begin{document}

\verso{Preprint Submitted for review}

\begin{frontmatter}

\title{Computer Vision Model Compression Techniques for Embedded Systems: A Survey}

\author[1,2]{Alexandre \snm{Lopes}\corref{cor1}}
\cortext[cor1]{Corresponding author:}
\emailauthor{alexandre.ribeiro@venturus.org.br}{Alexandre Ribeiro Lopes}

\author[1]{Fernando Pereira \snm{dos Santos}} 
\author[1]{Diulhio \snm{de Oliveira}}
\author[1]{Mauricio \snm{Schiezaro}}
\author[2]{Helio \snm{Pedrini}}

\address[1]{ Venturus - Innovation \& Technology , Campinas, Brazil, 13086-530}
\address[2]{Institute of Computing, University of Campinas, Campinas, Brazil, 13083-852}

\received{\today}

\begin{abstract}
Deep neural networks have consistently represented the state of the art in most computer vision problems. In these scenarios, larger and more complex models have demonstrated superior performance to smaller architectures, especially when trained with plenty of representative data. With the recent adoption of Vision Transformer (ViT) based architectures and advanced Convolutional Neural Networks (CNNs), the total number of parameters of leading backbone architectures increased from 62M parameters in 2012 with AlexNet to 7B parameters in 2024 with AIM-7B. Consequently, deploying such deep architectures faces challenges in environments with processing and runtime constraints, particularly in embedded systems. This paper covers the main model compression techniques applied for computer vision tasks, enabling modern models to be used in embedded systems. We present the characteristics of compression subareas, compare different approaches, and discuss how to choose the best technique and expected variations when analyzing it on various embedded devices. We also share codes to assist researchers and new practitioners in overcoming initial implementation challenges for each subarea and present trends for Model Compression. 
\end{abstract}


    \end{frontmatter}

\linenumbers

\section{Introduction}
\label{secIntroduction}


Deep learning methods have achieved outstanding results in various computer vision applications, including object detection~\cite{dai2021general}, face recognition~\cite{guo2019survey}, emotion detection~\cite{saxena2020emotion}, and video surveillance for crowd analysis~\cite{sreenu2019intelligent}. These methods are favored by increased computing power, enabling models to scale to over a billion parameters~\cite{he2016deep,szegedy2016rethinking,liu2022convnet}. Cloud computing provides capable resources for processing large-scale deep learning models but requires internet connectivity. Hence, it is unsuitable for many Internet of Things (IoT) applications needing more stable high-speed connections, traffic, robust data transfer rate, and low-latency response~\cite{branco2019machine}. In such cases, end-device processing arises in many situations to meet real-time requirements. End-devices, such as embedded systems and mobile devices, usually have limited processing power, RAM, storage, and power consumption, which makes them impractical for handling large deep learning models. Efficient neural network architectures, such as the MobileNets~\cite{howard2017mobilenets,sandler2018mobilenetv2} and RealTimeStereo~\cite{Chang_2020_ACCV}, were proposed to overcome those requirements.

\begin{figure*}[!htb]
\centering
\includegraphics[width=0.85\linewidth]
{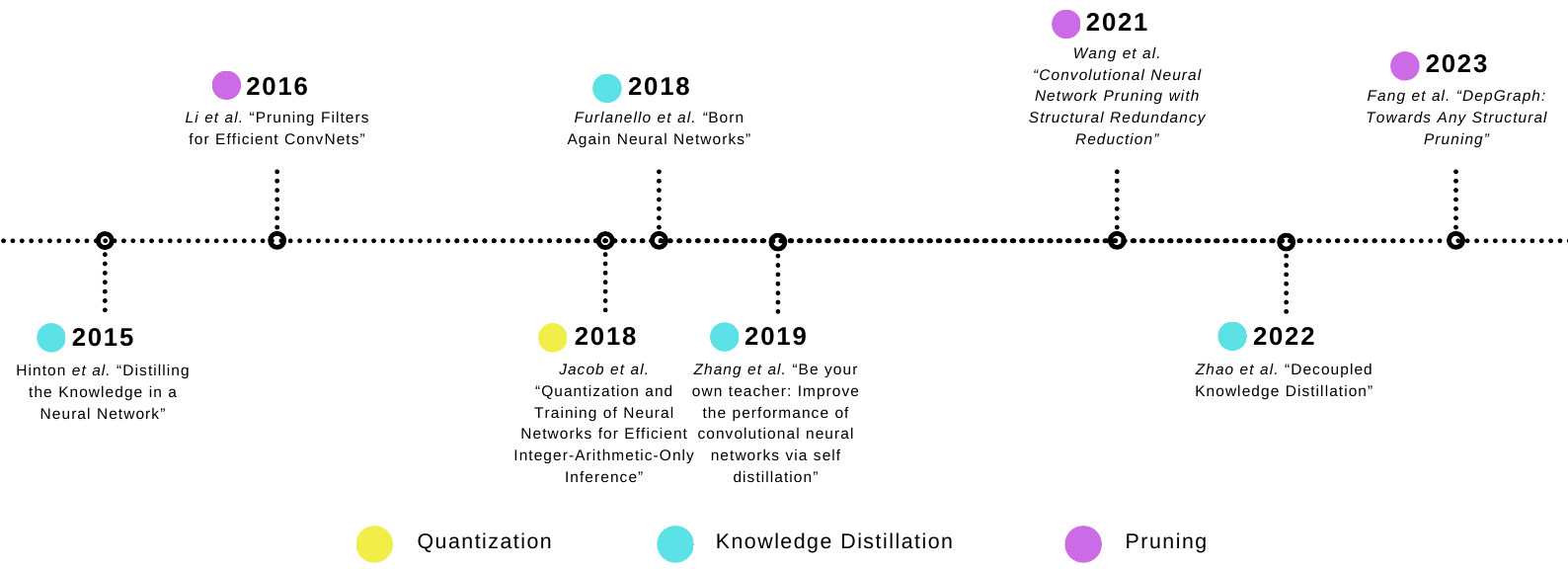}
\
\vspace*{-1.5mm}
\caption{Most influential proposed techniques since 2015 for Model Compression applied for Computer Vision. Low-Rank Factorization was omitted here due to its recent lack of usage in the field. A combination of the number of citations and novelty in the paper determined the most influential papers.}
\label{fig:grap}
\end{figure*}

Despite significantly reducing computational power, these small architectures present lower application metrics, which can be determinant for deploying an application in real-world scenarios. Naturally, more complex models tend to perform better if the training data is representative and in adequate quantity~\cite{Ribani2019survey,ponti2021training}. Instead of proposing smaller architectures with the same training procedures as complex networks, model compression strategies aim to reduce or modify large networks. Compression techniques allow powerful models to reduce execution time and memory consumption. Several studies have experimentally proven that compression techniques yield better results than conventional training in more simplified models~\cite{liu2018rethinking}. These techniques leverage methods such as Knowledge Distillation~\cite{dai2021general,liu2019structured,xu2018pad,qin2021efficient}, Network Pruning~\cite{wang2021recent,wang2021convolutional,yu2018nisp}, Network Quantization \cite{courbariaux2015binaryconnect,rastegari2016xnor, krishnamoorthi2018quantizing,jacob2018quantization, wang2019haq}, or Low-Rank Matrix Factorization \cite{sainath2013low,denil2013predicting,yu2017compressing,li2018constrained}. Therefore, understanding these techniques is crucial for applications that require limited computational power.

Despite recent efforts in the literature to survey and categorize model compression techniques applied across various artificial intelligence fields~\cite{cheng2017survey,li2023model,choudhary2020comprehensive,nan2019deep,berthelier2021deep}, some works focus their survey on specific applications, such as on Natural Language Processing (NLP)~\cite{gupta2022compression,wang2024model} or Computer Vision~\cite{goel2020survey}, due to specifications that emerge in each application area. For instance, object detection involves specific adaptations that are not shared with NLP tasks.

Our work aims to update previous literature surveys applied to Computer Vision tasks with recent papers and present a novel discussion of performance comparison between techniques of distinct compression categories over different devices. We also discuss trends in each compression subcategory and present a repository with examples that combine different approaches, allowing researchers and new practitioners to understand the domain more easily.

Our literature review enabled us to identify and recommend the most influential and novel papers on compression in the past ten years, providing an essential reading list for beginners, as shown in Figure~\ref{fig:grap}. It is important to note that this figure does not include any Low-Rank Matrix Factorization papers, primarily due to their almost nonexistent application in Computer Vision for model compression.

The remainder of this paper is structured as follows. In the next section, we discuss the methodology used to perform this paper's literature review and inclusion criteria. Section~\ref{secCompression} categorizes the compression techniques into four subareas: Knowledge Distillation, Network Pruning, Network Quantization, and Low-Rank Matrix Factorization. In Section~\ref{secMeasuresData}, we discuss metrics applied for different tasks and the datasets used to benchmark compression techniques. Section~\ref{perf} presents the performance comparison between the most popular datasets. In Section~\ref{disc}, we extensively discuss how to compare different methods, the effects of compression on various devices, trends for model compression, and the most prominent subareas. In Section~\ref{caseStudies}, we present the prepared case study with example codes. Finally, in Section~\ref{secConclusion}, we present conclusions for this survey.

\section{Methodology}
\label{sec:met}
We conducted an extensive literature review of compression techniques focused on Computer Vision, utilizing previous surveys as foundational references~\cite{cheng2017survey,li2023model,choudhary2020comprehensive,nan2019deep,berthelier2021deep,goel2020survey}. We filtered papers using the name of the compression area (Knowledge Distillation, Pruning, Quantization, or Low-Rank Matrix Factorization), followed by the ``Computer Vision'' keyword or a list of fourteen keywords of the most published computer vision subtasks, such as semantic segmentation, image classification, object detection, depth estimation, and action recognition.

\begin{figure*}[!htb]
  \centering
\tikzset{every picture/.style={line width=0.75pt}} 

\begin{tikzpicture}[x=0.75pt,y=0.75pt,yscale=-1,xscale=1]

\draw   (179,177) .. controls (179,172.58) and (182.58,169) .. (187,169) -- (261,169) .. controls (265.42,169) and (269,172.58) .. (269,177) -- (269,201) .. controls (269,205.42) and (265.42,209) .. (261,209) -- (187,209) .. controls (182.58,209) and (179,205.42) .. (179,201) -- cycle ;
\draw   (129,288.83) .. controls (129,284.42) and (132.58,280.83) .. (137,280.83) -- (211,280.83) .. controls (215.42,280.83) and (219,284.42) .. (219,288.83) -- (219,312.83) .. controls (219,317.25) and (215.42,320.83) .. (211,320.83) -- (137,320.83) .. controls (132.58,320.83) and (129,317.25) .. (129,312.83) -- cycle ;
\draw   (229,288.8) .. controls (229,284.38) and (232.58,280.8) .. (237,280.8) -- (311,280.8) .. controls (315.42,280.8) and (319,284.38) .. (319,288.8) -- (319,312.8) .. controls (319,317.22) and (315.42,320.8) .. (311,320.8) -- (237,320.8) .. controls (232.58,320.8) and (229,317.22) .. (229,312.8) -- cycle ;
\draw   (329,288.8) .. controls (329,284.38) and (332.58,280.8) .. (337,280.8) -- (411,280.8) .. controls (415.42,280.8) and (419,284.38) .. (419,288.8) -- (419,312.8) .. controls (419,317.22) and (415.42,320.8) .. (411,320.8) -- (337,320.8) .. controls (332.58,320.8) and (329,317.22) .. (329,312.8) -- cycle ;
\draw   (29,288.8) .. controls (29,284.38) and (32.58,280.8) .. (37,280.8) -- (111,280.8) .. controls (115.42,280.8) and (119,284.38) .. (119,288.8) -- (119,312.8) .. controls (119,317.22) and (115.42,320.8) .. (111,320.8) -- (37,320.8) .. controls (32.58,320.8) and (29,317.22) .. (29,312.8) -- cycle ;
\draw    (74,231.6) -- (73.89,275.75) -- (73.88,276.75) ;
\draw [shift={(73.88,279.75)}, rotate = 270.15] [fill={rgb, 255:red, 0; green, 0; blue, 0 }  ][line width=0.08]  [draw opacity=0] (7.14,-3.43) -- (0,0) -- (7.14,3.43) -- cycle    ;
\draw    (73.5,231.6) -- (373.5,231.6) ;
\draw    (224,208.77) -- (224,231.6) ;
\draw    (174,231.6) -- (173.89,275.75) -- (173.88,276.75) ;
\draw [shift={(173.88,279.75)}, rotate = 270.15] [fill={rgb, 255:red, 0; green, 0; blue, 0 }  ][line width=0.08]  [draw opacity=0] (7.14,-3.43) -- (0,0) -- (7.14,3.43) -- cycle    ;
\draw    (274,231.6) -- (273.89,275.75) -- (273.88,276.75) ;
\draw [shift={(273.88,279.75)}, rotate = 270.15] [fill={rgb, 255:red, 0; green, 0; blue, 0 }  ][line width=0.08]  [draw opacity=0] (7.14,-3.43) -- (0,0) -- (7.14,3.43) -- cycle    ;
\draw    (374,231.6) -- (373.89,275.75) -- (373.88,276.75) ;
\draw [shift={(373.88,279.75)}, rotate = 270.15] [fill={rgb, 255:red, 0; green, 0; blue, 0 }  ][line width=0.08]  [draw opacity=0] (7.14,-3.43) -- (0,0) -- (7.14,3.43) -- cycle    ;

\draw (44,287.67) node [anchor=north west][inner sep=0.75pt]  [font=\small] [align=left] {\begin{minipage}[lt]{43.64pt}\setlength\topsep{0pt}
\begin{center}
{\small Knowledge}\\{\small Distillation}
\end{center}

\end{minipage}};
\draw (138.67,296.33) node [anchor=north west][inner sep=0.75pt]  [font=\small] [align=left] {\begin{minipage}[lt]{49.16pt}\setlength\topsep{0pt}
\begin{center}
{\small Quantization}
\end{center}

\end{minipage}};
\draw (251,296.33) node [anchor=north west][inner sep=0.75pt]  [font=\small] [align=left] {\begin{minipage}[lt]{34.34pt}\setlength\topsep{0pt}
\begin{center}
Pruning
\end{center}

\end{minipage}};
\draw (336.33,287.33) node [anchor=north west][inner sep=0.75pt]  [font=\small] [align=left] {\begin{minipage}[lt]{55.25pt}\setlength\topsep{0pt}
\begin{center}
Low-Rank\\Factorization
\end{center}

\end{minipage}};
\draw (182.67,176) node [anchor=north west][inner sep=0.75pt]  [font=\small] [align=left] {\begin{minipage}[lt]{56.78pt}\setlength\topsep{0pt}
\begin{center}
Model \\Compression
\end{center}

\end{minipage}};
\draw (9.1,246) node [anchor=north west][inner sep=0.75pt]  [font=\Large] [align=left] {{\small 1912 papers}};
\draw (109.1,246) node [anchor=north west][inner sep=0.75pt]  [font=\Large] [align=left] {{\small 1458 papers}};
\draw (208.6,246) node [anchor=north west][inner sep=0.75pt]  [font=\Large] [align=left] {{\small 1383 papers}};
\draw (311.6,246) node [anchor=north west][inner sep=0.75pt]  [font=\Large] [align=left] {{\small  \ 262 papers}};

\end{tikzpicture}
  \caption{Model Compression Technique subdivision. We categorized Model Compression papers into four different areas. We also sort them based on Computer Vision papers, including each subcategory quantity of papers found from Jan/2021 to Mar/2024.} 
  \label{fig:subd}
\end{figure*}
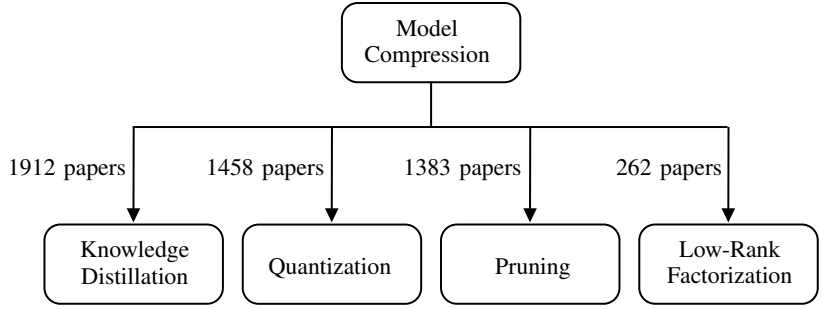

The literature review was carried out using Google Scholar and Scopus, resulting in a substantial selection of papers. For instance, Scopus provided 4,765 papers only from January 2021 to March 2024 (see Figure~\ref{fig:subd}). Due to the elevated number of papers, we used the following inclusion criteria: novelty, relation to model compression proposals, state-of-the-art results, and number of citations. A novel approach indicates that the paper presents new insights or techniques that significantly advance efficiency or performance, and we assigned grades from 1 to 5 to the papers based on this criterion. We only included papers with a score greater than 3. The number of citations criterion was not applied to papers from 2024. We excluded papers with 24 or fewer citations if they were from 2021 or earlier and those with 9 or fewer citations if they were from 2022 or 2023. For example, we discarded 1,532 (80.1\%) Knowledge Distillation papers from Scopus. We also excluded 262 papers related to Low-Rank Factorization because they were unrelated to model compression proposals. Consequently, the analysis of papers in this category in the following sections is limited. This survey also discusses papers with higher novelty rankings and highly cited papers in greater detail.


\section{Compression Techniques}
\label{secCompression}

In general, \textbf{Knowledge Distillation} aims to equate the output of a more compact network, called student, with a more complex one, called teacher. Thus, the student learns to simulate the outputs from this teacher, adjusting the weights according to this learning. \textbf{Network Pruning}, on the other hand, aims to eliminate redundant elements contained in the structure, reducing its size and inference time. \textbf{Network Quantization} modifies the weights with values from floating point to low-bit and can also be applied along with pruning after validating the redundant elements. Moreover, the idea of \textbf{Low-Rank Matrix Factorization} is to decompose the parameters from the architecture through a matrix tensor.

\subsection{Knowledge Distillation}
\label{subDistillation}

\begin{figure}[!htb]
\centering
\includegraphics[width=0.85\linewidth]{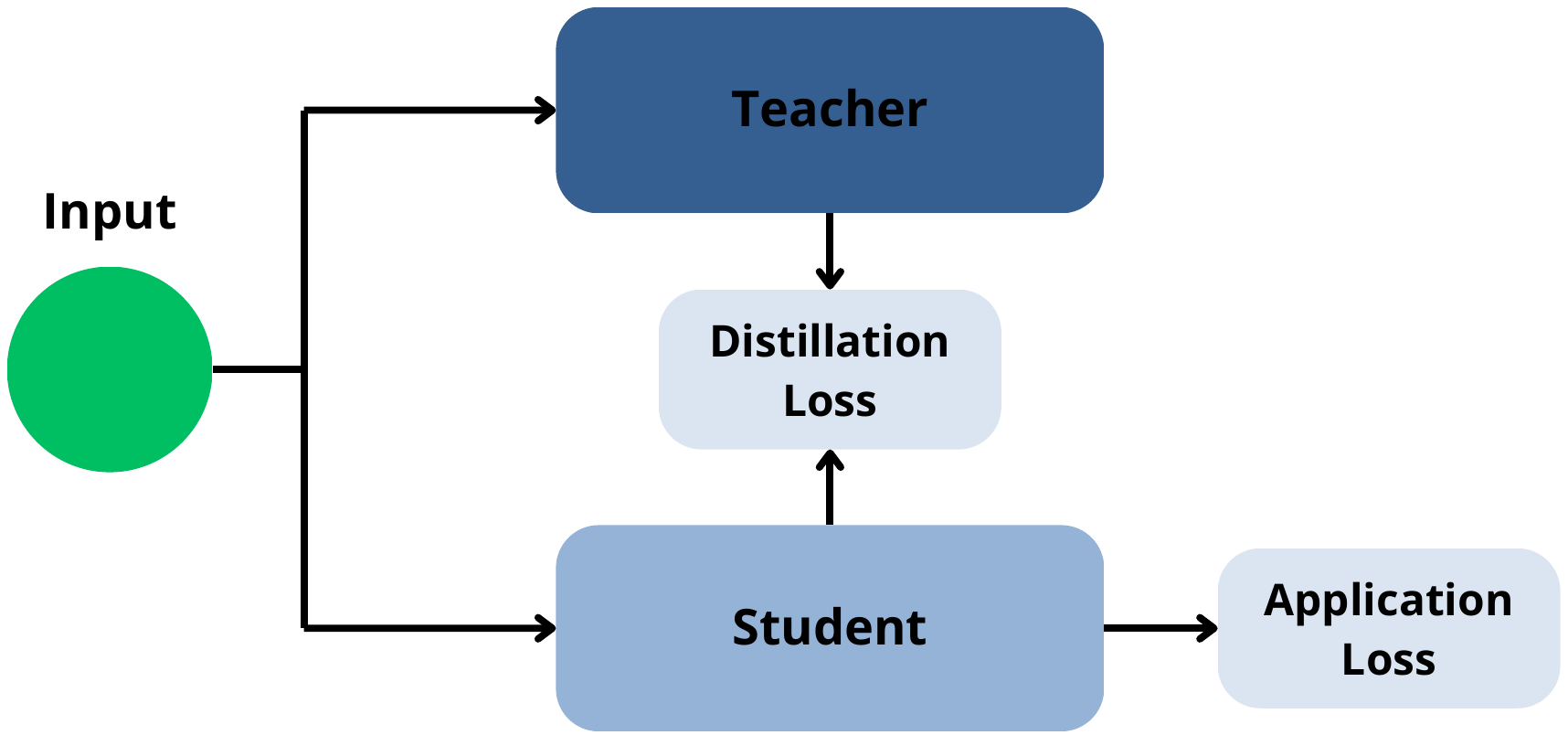}
\vspace*{-1mm}
\caption{Knowledge Distillation General Schematic. The teacher model usually receives the same input data as the student, and both features form the distillation loss, where the student will learn to mimic the teacher. The student can also have an application-dependent loss that varies depending on the application, such as cross-entropy for classification problems.}
\label{fig:distillation}
\end{figure}

The Knowledge Distillation (KD) process involves transferring the knowledge of a source network (teacher model) to a distilled network (student model)~\cite{hinton2015distilling}, as illustrated in Figure~\ref{fig:distillation}. The teacher model is usually a larger, pre-trained model that transfers knowledge to the student model during training. This process allows better results for the student architecture since its training will be molded to learn the behavior of the teacher model. Initially, the KD process involved image classification problems, but it was extended for almost all computer vision applications, such as object detection~\cite{dai2021general}, semantic segmentation~\cite{liu2019structured}, depth estimation~\cite{xu2018pad}, and medical imaging~\cite{qin2021efficient}.


Hinton~\etal~\cite{hinton2015distilling} proposed an $\mathcal{L}_{\text{KD}}$ loss term to mimic the teacher's model output (Equation~\ref{eq:1}). The $\mathcal{L}_{\text{KD}}$ is the Cross-Entropy (CE) of the probability distributions predicted by the softmax ($\sigma$) over the logits of the teacher ($z_t$) and the student ($z_s$) with a variable term called Temperature ($T$). The Temperature, $T>0$, controls how soft the probability distribution over the classes will be. This term is usually multiplied by a $T^2$ optional term that accounts for the $\frac{1}{T^2}$ term that appears in the cross-entropy gradient, as seen in Equation~\ref{eq:2}. The cross-entropy loss usually is substituted by the Kullback–Leibler (KL) divergence~\cite{aguilar2020knowledge,huang2022knowledge,tian2021knowledge} (Equation~\ref{eq:3}), since the KL divergence between the probability distributions $P$ and $Q$ is the same as the cross-entropy (CE) with the addition of the entropy of the $P$ distribution ($\text{KL}(P,Q) = \text{E}(P) + \text{CE}(P,Q)$). Additionally, Additionally, the KL divergence goes to zero if the distributions are equal, while CE does not guarantee a minimum value of zero. Equation~\ref{eq:3} can also be interpreted as Equation~\ref{eq:1} with the addition of the entropy of the teacher probability distribution. Since the teacher is a pretrained network, this value can be interpreted as a constant.

The $\mathcal{L}_{\text{KD}}$ loss is usually combined with an application-dependent loss term, which for image classification can be determined by the cross-entropy $\mathcal{L}_{\text{CE}}$ (Equation~\ref{eq:4}) or any desired loss function that produces the final loss $\mathcal{L}$ (Equation~\ref{eq:5}). In Equation~\ref{eq:5}, the $\alpha \in (0,1]$ term controls a balance between the importance of the cross-entropy term $\mathcal{L}_{\text{CE}}$ and the $\mathcal{L}_{\text{KD}}$ terms to the final loss. Therefore, we can control how much contribution the teacher will impose on the student and how much the student will learn directly from the training distribution. Knowledge Distillation is also achieved using other objective functions than KL divergence and cross-entropy, such as the Mean Squared Error (MSE) from the logits $z_t$ and $z_s$~\cite{kim2021comparing, wang2019distilling}.
\begin{equation} \label{eq:1}
\mathcal{L}_{\text{KD}} = -\sum_{i} \sigma_i (\frac{z_t}{T}) \times \log \sigma_i (\frac{z_s}{T})
\end{equation}
\begin{equation} \label{eq:2}
\frac{\partial \mathcal{L}_{\text{KD}}}{\partial z_s} \approx  \frac{1}{T^2} \left ( \sigma \left (  z_s \right ) - \sigma \left ( z_t \right ) \right )
\end{equation}
\begin{equation} \label{eq:3}
\mathcal{L}_{{\text{KD}}_{\text{KL}}} = \sum_{i} \sigma_i (\frac{z_t}{T}) \times \left (\log \sigma_i (\frac{z_t}{T}) -\log \sigma_i (\frac{z_s}{T})  \right ) 
\end{equation}
\begin{equation} \label{eq:4}
\mathcal{L}_{\text{CE}} = -\sum_{i}y_i \times \log \sigma_i (z_s)
\end{equation}
\begin{equation} \label{eq:5}
\mathcal{L}  = \alpha \mathcal{L}_{\text{KD}} + (1-\alpha) \mathcal{L}_{\text{CE}}
\end{equation}

The Knowledge Distillation process can be divided into three groups of strategies: offline, online, and self-distillation.

\paragraph{\textbf{Offline Distillation}} The training can be divided into two stages in offline distillation. First, the teacher model is solely trained. In the next stage, the teacher model weights are frozen, and the teacher will be used in the process of the student model training in the terms presented in Equation~\ref{eq:3}, where the results produced by the teacher's network for the training data is used to help the student. Therefore, the teacher model will distill its previously acquired knowledge to the student.

\paragraph{\textbf{Online Distillation}} Here, the training process is simplified to have a one-stage process where the weights of the teacher are also trainable during the student's training stage, and no pre-trained teacher is required. This simultaneous distillation process usually does not rely on having a larger separate teacher model to distill its knowledge to a smaller one. Instead, it explores strategies such as using an ensemble of the same network that will jointly learn collaboratively and teach each other during the training data~\cite{zhang2018deep}, having a teacher model formed by an ensemble of the last blocks from the student network~\cite{zhu2018knowledge}, or by using codistillation~\cite{anil2018large}, a process where two or more networks train on disjoint subsets of the training data and share knowledge during the training process.

\paragraph{\textbf{Self-Distillation}} The self-distillation process involves distilling sections or the logits of the teacher network to a student that presents the same architecture as the teacher. Unlike Online Distillation, self-distillation does not involve jointly learning with the teacher during the student training phase. The first proposed strategy divides a network into four sections and adds independent classifiers in each section composed of a bottleneck, fully connected, and softmax layers~\cite{zhang2019your}. This process creates a network with four different outputs divided into sections. Then, the successor section distills each predecessor section. The extra bottlenecks, fully connected, and softmax layers are discarded in the final network. Other strategies explore self-distillation in distinct forms, trying to distill knowledge from networks that share the same weights and are inputted with images from the same class~\cite{yun2020regularizing}, or intra-stage distillation combined with inter-stage distillation~\cite{gou2023multi}.


Despite having different distillation strategies regarding the teacher-student training process, KD also explores what sort of information from the teacher is learned by the student. In this condition, we have three categories~\cite{gou2021knowledge}: Response-Based Knowledge, Feature-Based Knowledge, and Relation-Based Knowledge.

\paragraph{\textbf{Response-based Knowledge}} In this strategy, the objective is to match the logits produced by the teacher and student, i.e., the outputs of the last layer of the teacher and student will be evaluated in the loss function, as in Equation~\ref{eq:1}. Since it is the simplest form of matching, vanilla strategies of KD can be categorized as response-based techniques~\cite{hinton2015distilling,kim2016sequence}. This knowledge scheme is mainly applied for classification tasks, although having applications in different areas, such as object detection~\cite{xing2022dd} and transfer-domain between RGB and depth~\cite{gupta2016cross}. Some applications present an increased complexity of the output structures and disfavor the KD only applied to the logits, e.g., object detection networks generally predict at least a grid output with width, height, $x$, $y$, and class probability for each image. Zhao~\etal~\cite{zhao2022decoupled} proposed a strategy that breaks the classical KD into target and non-target classes, allowing the user to control the contribution of each for the KD.

\paragraph{\textbf{Feature-based Knowledge}} In addition to matching the logits of the student and teacher models, many works explore the relation between the intermediate features of the teacher and student~\cite{dai2021general,li2017mimicking,de2022structural}. The intuition is that it is easier to produce the same distribution in the output if the student network's internal blocks produce similar feature maps compared to the teacher network. Therefore, adding a term to match intermediate features could improve the model results, favoring complex output scenarios, such as object detection. Seminal investigations in KD applied to object detection~\cite{chen2017learning} adapt intermediate layers to match features from the teacher and student. Recently, Chang~\etal~\cite{chang2023detrdistill} proposed a general-form KD strategy using Transformer-based architectures that includes several decoder stages matching student and teacher internal features.

\paragraph{\textbf{Relation-based Knowledge}} Differently from the two previous categories, relation-based knowledge aims to transfer structural relations of features instead of transferring features by themselves from the teacher to the student~\cite{park2019relational}. Therefore, a function $\psi$ will be applied to the student and teacher's features to transform the feature space and then match both to minimize the difference of feature maps in the new feature space. Different approaches have been proposed, including an attention-guided distillation that finds an attention match between features of multiple stages of the student and teacher networks~\cite{zhang2020improve} and an attention distillation but matching all possible candidates of feature maps from the teacher to the student~\cite{ji2021show}.
\newline

A pseudocode for a general Knowledge Distillation algorithm applied to image classification problems can be seen in Algorithm~\ref{alg:kd}. Steps~1 and~2 initialize the student model with random weights and load the weights of the teacher model, respectively. After performing the forward pass of both models (``forwardPass'' function), the $z^t$ and $z^s$ values can represent the logits of the models in response-based knowledge or the features of distinct stages of the teacher and student models for feature-based knowledge. For relation-based knowledge, both ``forwardPass'' steps should also include the function $\psi$ to transform both feature spaces. The Knowledge Distillation Loss (``KDLoss'' function) is usually calculated using either the $\mathcal{L}_{\text{KD}}$ (Equation~\ref{eq:1}) or the $\mathcal{L}_{\text{KD}_\text{KL}}$ (Equation~\ref{eq:3}), despite other possibilities such as the MSE. The desired classification loss is calculated in Step~7, and, finally, the final loss is composed of the knowledge distillation and classification terms used to update the weights of the student.

\begin{algorithm}
	\small
	\caption{- \textit{Knowledge Distillation} - Conventional Pipeline}
	\label{alg:kd}
	\begin{algorithmic}[1]
		\Require
		\Statex $M^t$: Deep Network Teacher Network
		\Statex $M^s$: Deep Network Student Network
		\Statex $W^s$: Parameters of $M^s$
		\Statex $X$: Data training
        \Statex $Y$: Data training labels
        \Statex $T$: Temperature Parameter
        \Statex $\alpha$: Knowledge Distillation Control Parameter

		\State $W^s \longleftarrow \text{initialize}(M^s)$
		\State $M^t \longleftarrow \text{loadPretrained}(M^t)$		

		\While{$stop-criterion$}
			\State ${z}^{s} \longleftarrow \text{forwardPass}(M_s, X)$
			\State ${z}^{t} \longleftarrow \text{forwardPass}(M_t, X)$

			\State $\mathcal{L}_{\text{KD}} \longleftarrow \text{KDLoss}(z^t, z^s, T)$

			\State $\mathcal{L}_{\text{C}} \longleftarrow \text{ClassificationLoss}(z^s, Y)$

			\State $\mathcal{L} \longleftarrow \alpha \mathcal{L}_{\text{KD}} + (1-\alpha) \mathcal{L}_{\text{C}}$

            \State $W_s \longleftarrow \text{updateWeights}(\mathcal{L})$

		\EndWhile
	\end{algorithmic}
\end{algorithm}

\subsection{Network Pruning}
\label{subPrunning}

\begin{figure}
\begin{tabular}{cc}

\includegraphics[height=0.26\columnwidth]{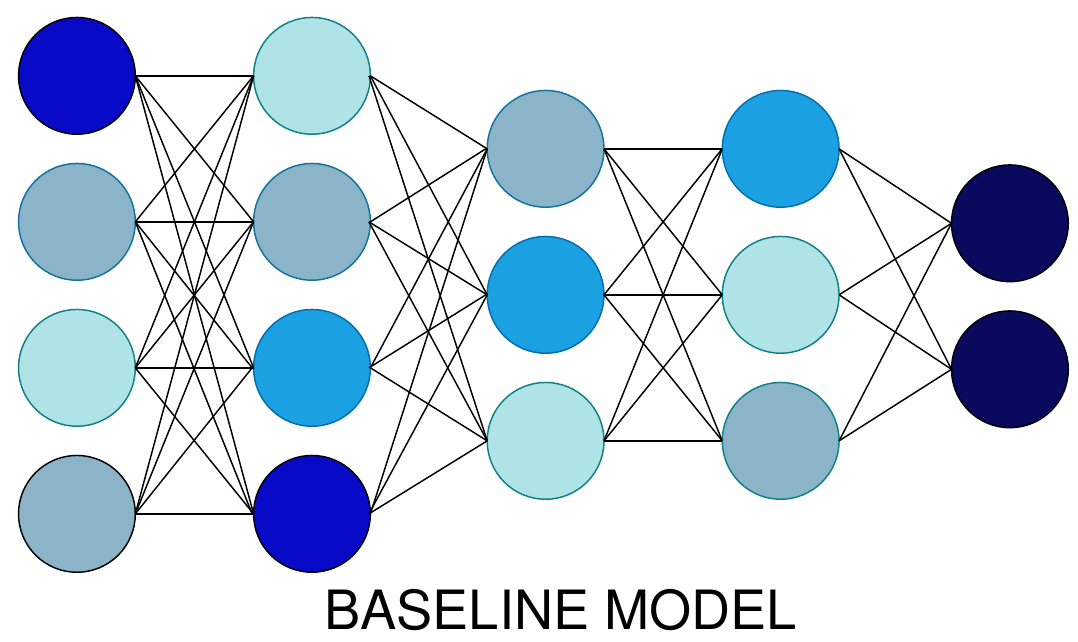}
  &
\includegraphics[height=0.26\columnwidth]{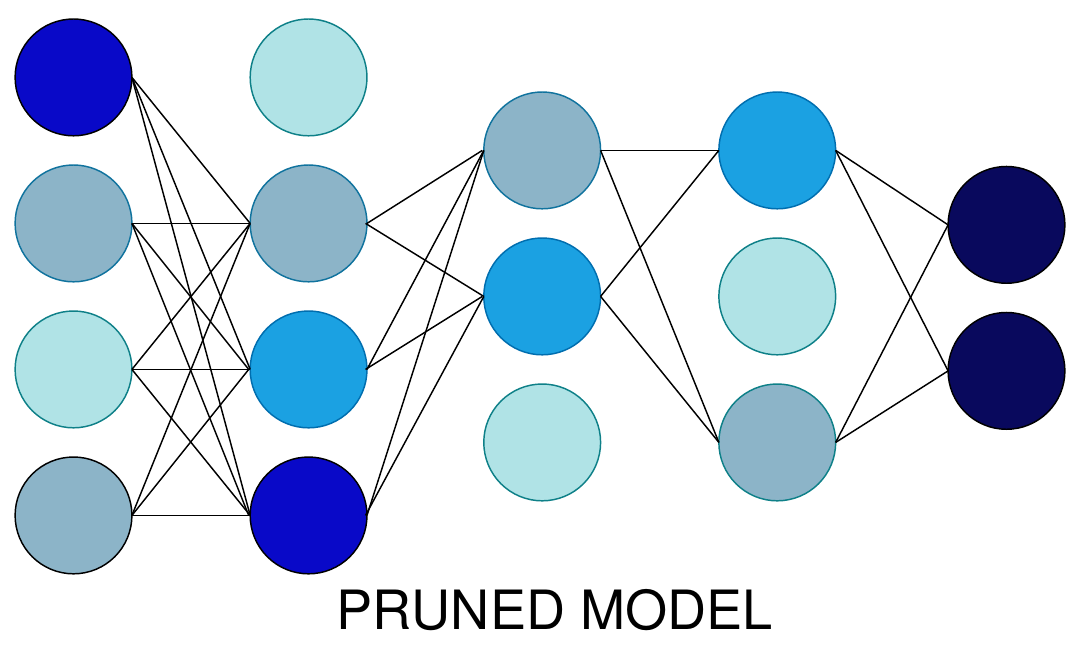}

   \\             
 \end{tabular}
\vspace*{-1mm}  
\caption{Network Pruning. The left side shows a binary model trained by conventional techniques and/or advanced strategies comprising five processing layers. The right side shows the same model after pruning. Different shades of blue indicate the degree of relevance of each element, determined by a rule: the darker the shade, the more important the element. Hence, connections to and from the low-relevance elements are removed after establishing a pruning approach. Irrelevant parameters from input and output layers remain after pruning. Then, the pruned structure can be fine-tuned. The pruned elements can be filters, channels, or structures.}

\label{fig:pruning1}

\end{figure}

Network Pruning is a model compression technique based on discarding low-relevance parameters for prediction. Consequently, the neural network becomes smaller and faster in its execution, being able to be directly applied in systems with hardware restrictions~\cite{wang2021recent}. First, the usual three-stage procedure, as illustrated in Figure~\ref{fig:pruning1}, is to consider a complex network to be trained from scratch or a pre-trained model that is available~\cite{he2016deep,szegedy2016rethinking,liu2022convnet}. Afterward, as a second step, a pruning technique is applied to adjust the structure to the desired size and, subsequently, adjust the weights to the target domain with fine-tuning~\cite{liu2018rethinking}. This standard procedure is based on two concepts. The first one is that previously deep models trained on representative data  provide good generalization even with the removal of redundant parameters. The second fact is that the remaining parameters are essential for the efficiency of the resulting model~\cite{liu2018rethinking,yu2018nisp}. However, this traditional pipeline has already undergone some changes, such as removing some elements and discarding all weights to perform a new training from scratch, considering a pre-trained structure~\cite{liu2018darts} or even initialized with random weights~\cite{wang2020pruning}.

When deciding to prune a neural network, four aspects~\cite{wang2021recent} must be considered in the pruning function: what strategy to use, how many connections to prune, what to prune, and how to structure the process. \textbf{What strategy to use} is directly related to the sparsity of the structure. If only single weight elements are pruned, we call it unstructured pruning. Generally these are zeroed to maintain the structure of the network. On the contrary, if more than single elements are pruned, the process is categorized as structured pruning. This process is executed over different levels of granularity, i.e., prune can completely remove filters, channels, or even more complex structures in the network.

Most studies apply structured pruning to offer more convenience for hardware constraints~\cite{wang2021recent}. \textbf{How many connections to prune} can be decided in two ways: determine an exact amount, which can be characterized globally, considering the entire architecture, or layer-wise, considering an amount per layer. After that, the most critical aspect of implementing Network Pruning is properly identifying \textbf{what to prune}, comprising three main alternatives~\cite{wang2021convolutional}, as described as follows.

\paragraph{\textbf{Ranking and Pruning Parameters}} It applies some criterion to select the relevant elements. The evaluation can be carried out considering simpler arithmetic functions, such as the sum of the absolute values of the weights~\cite{li2016pruning} or the percentage of weights with values equal to zero~\cite{hu2016network}, to more complex calculations, such as verification of the inter-channel relationship~\cite{peng2019collaborative} or structural redundancy~\cite{wang2021convolutional}. 

\paragraph{\textbf{Reconstruction-based Methods}} It seeks to minimize the error in the reconstruction of feature maps by comparing the pruned model to the pre-trained model~\cite{yu2018nisp}. Suppose we have a deep network with $N$ layers and $k$ filters in each layer. Therefore, each of these $k_{N_i}$ filters is composed of different weights and provides a $c_{N_i}$ contribution to the final performance of its inference~\cite{wang2021convolutional}. The contribution can be measured directly by the difference in performance after pruning a filter $k_{N_i}$ or by changing the training error. Hence, if the filter $k_{N_i}$ is pruned and its $c_{N_i}$ contribution is less than a certain threshold, the pruning is well supported by the network. An irrelevant filter may be maintained during the reconstruction to reduce the pruned model error. This concept can be generalized in terms of channels and structures.
 
\paragraph{\textbf{Similarity Measurement}} It compares the various elements to each other to eliminate some of the most similar through geometric median~\cite{he2019filter} or clustering~\cite{zhou2018online}, for example. Clustering can be complex to implement because of the weight's variability. Therefore, the original loss function can be combined with a cluster loss during the training to facilitate the similarity. Recently, Fang~\etal~\cite{fang2023depgraph} proposed the DepGraph approach to create a fully automatic dependency graph to model the dependency between layers of the network and group those parameters for pruning. In this approach, the parameter importance is substituted to the analysis of the groups, determining which groups are the safest ones to be removed without substantially degrading the performance.
\newline

It is also essential to understand \textbf{how to structure the process}, which can be specified by three main alternatives: one-shot, directly applying the sparsity in a single step and, subsequently, fine-tuning directly; progressive, applying sparsity little by little during the initial training of the neural network; or iterative, applying the sparsity process repeatedly over a few cycles~\cite{wang2021recent}.

\begin{algorithm} 
    \small
    \caption{- \textit{Network Pruning} - Conventional Pipeline} 
    \label{alg:pipelinePruning}
    \begin{algorithmic}[1]
        \Require 
        \Statex $M$: Standard Deep Network structure
        \Statex $W$: Parameters of the $M$
        \Statex $X$: Data Training
        \Statex $i$: iteration

        \Comment{First step}
        \State $W_0 \longleftarrow initialize(M_0)$
        \State $W_0 \longleftarrow trainToConvergence(M_0, W_0, X)$
        \While{$stop-criterion$}
            \State $M_i, W_i \longleftarrow pruningRule(M_{i-1}, W_{i-1})$
            \Comment{Second step}
            \State $W_i \longleftarrow trainToConvergence(M_i, W_i, X)$
            \Comment{Third step}
        \EndWhile
    \end{algorithmic}
\end{algorithm}

A pseudocode based on~\cite{blalock2020state} that demonstrates the conventional pipeline for applying Network Pruning is available in Algorithm~\ref{alg:pipelinePruning}. The second and third steps can be performed iteratively, in which the pruning function is applied, the architecture is retrained, and the performance is evaluated, as performed in~\cite{ding2021resrep, wang2020neural, he2019filter}. The pruning function and training process can be repeated until the desired architecture size or a minimum performance threshold is reached. Other training strategies, such as the presented DepGraph, do not show iterative training modifying the structure of the network but can also be represented in this algorithm, as the ``prunningRule'' function in this case will be performed only once, and the ``trainToConvergence'' function do not alter the model.

\subsection{Network Quantization}
\label{subQuantization}

Network Quantization is a widely used method for neural network compression and acceleration, where the fundamental idea is converting the network parameters and inputs into a computationally lighter numerical representation. Training and inferring neural networks are computationally intensive tasks, especially when deep learning architectures are the main core~\cite{jacob2018quantization, wang2019haq}. Network parameters, such as weights, activation, and biases, are generally represented as 32-bit float-point numbers along the training and tuning phases. This numerical representation may be a problem for applications requiring embedded devices, where storage and computational performance limitations are present~\cite{gholami2021survey}. Furthermore, 32-bit float-point values require extensive mathematical operations on the same representation, implicating a lower power efficiency~\cite{wang2019haq}. Efficient numerical representations emerge as a particular solution, and as models are over-parameterized, there are opportunities to reduce the models’complexity with a low impact on the models’performance~\cite{gholami2021survey}. 

\begin{figure}[!htb]
\centering

\includegraphics[width=1\linewidth]{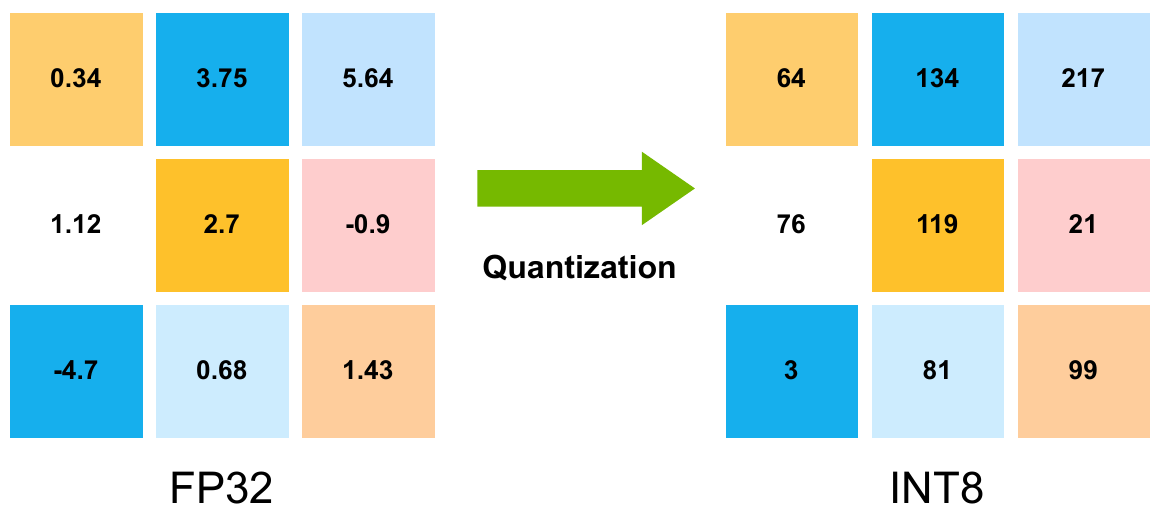}
\vspace*{-5mm}
\caption{Network Quantization. On the left, the original weights of a neural network are represented in a matrix format using 32-bit floating-point numbers. On the right, the network weights after the quantization process to an 8-bit integer.}
\label{fig:quantization}
\end{figure}

Instead of using 32-bit floating-point numbers, lower precision representations such as 8-bit integers~\cite{jacob2018quantization,wang2019haq} or even binary values are employed~\cite{courbariaux2015binaryconnect,rastegari2016xnor}. Figure~\ref{fig:quantization} illustrates the quantization process for a 32-bit floating-point neural network weights to an 8-bit integer. Doing so can lead to reduced memory footprint, lower latency, and better power efficiency of the neural network, enabling deployment on devices with hardware constraints~\cite{wang2019haq}.

While Network Quantization offers benefits regarding memory usage and inference speed, it also introduces a certain level of performance degradation~\cite{wang2019haq,gholami2021survey}. The reduction in precision can lead to a loss of information, which may result in decreased performance compared to the original network. This performance depends on factors such as the target precision, the specific network architecture, and the nature of the dataset being used~\cite{krishnamoorthi2018quantizing,gholami2021survey}.

When employing Network Quantization, balancing the trade-off between model size, computational efficiency, and accuracy is essential. One of the challenges is determining an appropriate precision level for each network parameter. Different types of layers or even individual weights may require different levels of precision to maintain a good trade-off between inference performance and memory/computational savings. Finding the right balance often involves experimentation and fine-tuning to achieve optimal results.

Regarding Network Quantization, two main techniques are commonly employed \cite{krishnamoorthi2018quantizing}, as described as follows.

\paragraph{\textbf{Quantization-Aware Training (QAT)}} Technique used to train neural networks with the knowledge that quantization will be applied later. The process involves simulating the effects of quantization during training, allowing the network to adapt and learn to minimize the performance degradation caused by reduced precision~\cite{mellempudi2017ternary, hubara2018quantized, choi2018pact, jacob2018quantization}. QAT typically involves introducing quantization steps during forward and backward passes, applying quantization-aware optimization algorithms, and fine-tuning the network to achieve optimal accuracy under the target precision constraints.

Algorithm \ref{alg:pipelineqat}, based on~\cite{jacob2018quantization}, demonstrates the conventional pipeline for applying Quantization-Aware Training. Note that this implementation reflects the concept of QAT; there may be differences compared to implementations in Deep Learning frameworks since these tools work in a graph-oriented manner. The basic idea of QAT is to work with a ``fake quantization" of the 32-bit floating point parameters in the forward step of the network. This is an exact simulation of the inference process. Afterward, the output is ``dequantized'' and used in the loss function. Finally, the error is used to compute the gradients and update the weights in 32-bit floating point format. It is important to note that the statistics of the weights are calculated at the beginning of the process, and they are essential in the quantization and ``dequantization'' process.

\begin{algorithm}
	\small
	\caption{- \textit{QAT} - Conventional Pipeline}
	\label{alg:pipelineqat}
	\begin{algorithmic}[1]
		\Require
		\Statex $M$: Standard Deep Network structure
		\Statex $W$: Parameters of the $M$
		\Statex $X$: Data training
        \Statex $Y$: Data training labels
        \Statex $\hat{Y}$: Deep Network outputs
		\State $W \longleftarrow initialize(M)$
				
		\While{$stop-criterion$}
			\State $S \longleftarrow computeStatistics(W)$
			\State $W_{q} \longleftarrow quantizeWeights(W, S)$
            \State $\hat{Y}_{q} \longleftarrow M(X, W_{q})$
            \State $\hat{Y} \longleftarrow desquantize(\hat{Y}_{q}, S)$
            \State $L \longleftarrow loss(Y, \hat{Y})$
            \State $W \longleftarrow updateWeights(L)$
		\EndWhile
	\end{algorithmic}
\end{algorithm}

\paragraph{\textbf{Post-Training Quantization} (PTQ)} An alternative approach where the quantization step is applied to a pre-trained neural network after the completion of regular training~\cite{gholami2021survey,nagel2019data, banner2019post, cai2020zeroq}. This technique converts the weights and activations of the network to lower precision representations. Post-training quantization can be a more straightforward approach than Quantization-Aware Training since it does not require modifying the training process. However, it may result in a slightly lower accuracy compared to QAT, as the network has not been explicitly trained to handle the effects of quantization. Nonetheless, post-training quantization remains popular for deploying quantized models in various applications.

Algorithm \ref{alg:pipelineptq} demonstrates the conventional pipeline for applying Post-Training Quantization. 
This method requires a calibration dataset passed through the network to extract statistics used for the numerical transformations within the quantization process.
Finally, the quantization is performed per layer, and the final model is rebuilt using the quantized weights.

\begin{algorithm}
\small
\caption{- \textit{PTQ} - Post-Training Quantization Pipeline}
\label{alg:pipelineptq}
\begin{algorithmic}[1]
\Require
\Statex $M$: Standard Deep Network structure
\Statex $W$: Parameters of the $M$
\Statex $X_{calib}$: Calibration data
\Statex $S$: Quantization scales and zero-points
\State $W \longleftarrow loadPretrainedModel(M)$
\State $S \longleftarrow computeCalibrationStatistics(W, X_{calib})$
	\State $W_{q} \longleftarrow quantizeWeights(W, S)$

	\For{$\forall l \in layers(M)$}
		\State $l_{q} \longleftarrow quantizeLayer(l, S)$
	\EndFor

	\State $M_{q} \longleftarrow rebuildModelWithQuantizedLayers(M, W_{q})$
	
\end{algorithmic}
\end{algorithm}

\subsection{Low-Rank Matrix Factorization}
\label{subMatrix}

Low-Rank Matrix Factorization (LRMF) is a technique employed in computer vision that aids in decomposing a given matrix into two or more matrices of lower rank. This approach approximates the original matrix by multiplying the lower rank matrices, thereby enabling the creation of a more condensed representation of the data and facilitating the identification of patterns within the content~\cite{sainath2013low,choudhary2020comprehensive,yu2017compressing}.

In this context, rank refers to the maximum number of linearly independent columns or rows of lower-rank matrices used to approximate the original matrix satisfactorily. LRMF offers several advantages, such as reducing network complexity and size in deep neural networks, which typically consist of multiple layers and many parameters. This makes it possible to create a more compact network, leading to lower memory requirements and decreased computational resource consumption~\cite{goel2020survey,yu2017compressing}.

Moreover, unlike many other techniques that are exclusively used during inference, LRMF can also be employed during training, which has the potential to expedite the training process and mitigate the issue of overfitting, as the lower rank matrices contain fewer parameters, effectively reducing the model's tendency to memorize noise or irrelevant information~\cite{choudhary2020comprehensive,goel2020survey}. 

Despite its advantages, LRMF also presents some challenges. One of the main difficulties lies in determining an appropriate rank for the factorization process.
The selection of an excessively low-rank matrix may result in information loss, which can decrease the model performance~\cite{li2018constrained,goel2020survey,yu2017compressing}. In contrast, selecting a high-rank matrix may mitigate the advantages of dimensionality reduction. Thus, the correct balance is crucial to the performance of LRMF techniques. For training, using LRMF may be a problem for large-scale datasets since the matrix decomposition is computationally expensive~\cite{goel2020survey}.
Additionally, architectures such as CNNs are designed to capture and learn spatial hierarchies through convolutional layers. Matrix factorization in CNNs involves decomposing the learned convolutional filters into lower-rank matrices. However, the spatial arrangement and interdependencies of the filters within the convolutional layers can be crucial for capturing meaningful features in the data. Applying LRMF without careful consideration may disrupt the spatial relationships encoded in the filters, potentially leading to loss of important information or degraded model performance \cite{goel2020survey}.

The techniques may effectively enhance the interpretability and generalization capabilities of the models. Moreover, the LRMF may reduce the memory and computational resources, allowing the deployment of more compact models on embedded and mobile devices. The LRMF reduces the parameters and may speed up between 30\% to 50\% the neural network latency~\cite{sainath2013low}, the memory requirements on 2 or 3 times for convolutional layers, and 5-13 times for dense layers~\cite{denil2013predicting}.

Algorithm \ref{alg:low_rank_matrix_factorization} presents the conventional pipeline for Low-Rank Matrix Factorization (LRMF) based on Singular Value Decomposition (SVD). This algorithm outlines the fundamental steps for reducing the dimensionality of a given matrix while preserving its essential structure.
In the LRMF pipeline, the input matrix $A$ is subjected to the Singular Value Decomposition (SVD), resulting in three matrices: $U$, $S$, and $V^{T}$. These matrices represent the orthogonal bases and singular values that capture the essential structure of $A$.
The algorithm then selects the top $k$ singular components to construct a low-rank approximation of the original matrix. This is achieved by retaining only the leading columns of $U$, the dominant elements on the diagonal of $S$, and the corresponding rows of the transpose of $V$.
With the reduced rank matrices, $U_{k}$, $S_{k}$, and $V^{T}_{k}$, the algorithm reconstructs an approximation of the original matrix $A$. This approximation aims to capture the most significant features of the input matrix while reducing its dimensionality.

\begin{algorithm}
    \small
    \caption{- \textit{LRMF} - Low-Rank Matrix Factorization}
    \label{alg:low_rank_matrix_factorization}
    \begin{algorithmic}[1]
        \Require
        \Statex $A$: Input matrix
        \Statex $k$: Desired rank

            \State $U, S, V^T \gets SVD(A)$
            \Comment{Apply Singular Value Decomposition}
            \newline
            \State Select the top $k$ singular components
            \State $U_k \gets$ \Call{FirstKColumns}{$U, k$}
            \State $S_k \gets$ \Call{FirstKDiagonal}{$S, k$}
            \State $V_k^T \gets$ \Call{FirstKRows}{$V^T, k$}
            \State $A_{approximated} \gets U_k \times S_k \times V_k^T$
            \Comment{Reconstruct the matrix with reduced rank}
    \end{algorithmic}
\end{algorithm}

\section{Metrics and Datasets}
\label{secMeasuresData}

The metrics of the model compression algorithms are generally based on the application they are developed for, such as Accuracy, Top-1 Accuracy, and Top-5 Accuracy for image classification~\cite{peng2019collaborative,ji2021show}, Average Precision for object detection~\cite{zhang2020improve,de2022structural}, and Dice Coefficient~\cite{qin2021efficient} for image segmentation. In order to assess the model compression improvements, some studies evaluate the number of parameters~\cite{kim2016sequence,li2017mimicking,zhang2020improve,chen2017learning}, inference time~\cite{li2017mimicking,zhang2020improve,chen2017learning}, Floating Point Operations per Second (FLOPs)~\cite{he2019filter, peng2019collaborative, wang2021convolutional}, and energy consumption~\cite{blakeney2020parallel}. 

Stanton~\etal~\cite{stanton2021does} proposed to investigate the fidelity of the KD training, i.e., the ability of a student network to match a teacher’s predictions. The fidelity comprises two complementary metrics: Average Top-1 Agreement and Average Predictive Kullback-Leibler (KL). The first is the Top-1 label agreement between the student and teacher networks, and the second is the KL divergence of the predicted teacher and student distributions.

The datasets used to compare compression techniques also vary based on the application they are developed for. Therefore, we will present further details on classification and detection problems since they are the most discussed problems for all compression subcategories.

Most proposed compression techniques compare their models in a classification configuration using at least one of the following datasets: ImageNet-1K~\cite{deng2009imagenet}, CIFAR-10, and CIFAR-100~\cite{krizhevsky2009learning}. Thus, these datasets appear as the most common benchmarking datasets for compression techniques. Despite that, highly cited papers in the field also present their results in other datasets, such as the Market-1501~\cite{zheng2015scalable}, ILSVRC-12~\cite{russakovsky2015imagenet}, Food101~\cite{kaur2017combining} and SUN397~\cite{xiao2010sun} datasets.

The CIFAR datasets are composed of 60,000 $32\times32$ RGB images. The CIFAR-10 includes ten distinct classes, while the CIFAR-100 presents images of 100 object classes. Both datasets are split into a training set of 50,000 images, a validation set of 5,000 images, and a test set of 5,000 images. The ImageNet-1K dataset contains 1.28 million training images and 50,000 validation images divided into 1,000 classes. The original images have a resolution of $469\times387$ pixels but are commonly resized to $224\times224$ pixels.

Papers that present object detection experiments or are solely focused on the detection problem rarely compare their results with other datasets than the MS COCO~\cite{lin2014microsoft} and PASCAL VOC~\cite{everingham2010pascal} datasets.

\begin{table*}[!ht]
\caption{Pruning and KD results for the CIFAR 10 dataset. The reported Baseline Model is the larger model used in the compression process. Since Pruning removes weights from the main network and does not involve different architectures, the baseline and compressed models share the same architecture. Results that are not available in the papers and could not be estimated are indicated by an em dash (---). \dag result not available in the paper and calculated to favor comparison. The table rows are sorted by papers’ published years}
\setlength{\tabcolsep}{5pt}
\centering
\label{tab:1}

\resizebox{\textwidth}{!}{%
\begin{tabular}{ c | c | c | c | c | c | c | c | c}


\hline

& \multirow{2}{4em} {\centering Method}  &  \multirow{2}{4em}{\centering Baseline Model} & \multirow{2}{6em}{\centering Compressed Model}  & \multirow{2}{6em}{\centering Params. Reduction (\%)} & \multirow{2}{6em}{\centering Baseline Model Acc.} & \multirow{2}{6em}{\centering Compressed Model Acc.} & \multirow{2}{4em}{\centering FLOPs Drop (\%)} & \multirow{2}{4em} {\centering Acc. Drop (\%)} \\
& & & & & & & \\
\hline 
 
\multirow{11}{*}{\rotatebox{90}{Pruning}} 


& Liu~\etal~\cite{liu2018rethinking} & DenseNet-BC-100 & DenseNet-BC-100 & 50 & 95.24 $\pm 0.17$ & 95.04 $\pm 0.15$ & --- & 0.20
\\


& Liu~\etal~\cite{liu2018rethinking} & VGG-19  & VGG-19 & 50 & 93.50 $\pm 0.11$ &  93.52 $\pm 0.10$ & --- & -0.02
 \\

& Liu~\etal~\cite{liu2018rethinking} & VGG-19  & VGG-19 & 95 & 93.50 $\pm 0.11$ &  93.34 $\pm 0.13$ & --- & 0.16 
\\

& NISP~\cite{yu2018nisp} & ResNet56  & ResNet56 & 42.6 & --- &  --- & 43.61 & 0.03 
\\


& Peng~\etal~\cite{peng2019collaborative} & ResNet56  & ResNet56 & --- & 93.50 &  93.69 & 47 & -0.19 
\\

& He~\etal~\cite{he2019filter} & ResNet56  & ResNet56 & --- & 93.59 &  93.49 & 52.6 & 0.10 
\\


& Wang~\etal~\cite{wang2020pruning} & ResNet56  & ResNet56 & 50 & 93.23 &  93.05 $\pm 0.19$ & --- & 0.18 
\\


& Wang~\etal~\cite{wang2020pruning} & VGG-19  & VGG-19 & 50 & 93.40 &  93.71 $\pm 0.08$ & --- & -0.31 
\\

& GReg-2~\cite{wang2020neural} & ResNet56  & ResNet56 & --- & 93.36 & 93.36 &  60.78 & 0 \\

& SRR-GR~\cite{wang2021convolutional} & ResNet56  & ResNet56 & --- & 93.38 &  93.75 & 53.8 & -0.37 
\\

& DepGraph~\cite{fang2023depgraph} & ResNet56  & ResNet56 & --- & 93.53 & 93.64 &  61.09 & -0.11 \\


\hline
\hline 

\multirow{2}{*}{\rotatebox{90}{KD}} & 

Xu~\etal~\cite{xu2023constructing}&  Wide-ResNet-28-4 & Wide-ResNet-16-2 & 88.21~\dag & 93.1 &  91.14 & --- & 1.96 \\ &

Xu~\etal~\cite{xu2023constructing} &  ResNet18 & VGG11
& 12.71~\dag & 93.2 &  89.12 & --- & 4.08 \\

\end{tabular}
}
\end{table*}

\section{Performance Comparison}
\label{perf}

The performance comparisons in the literature are segmented by subcategories. For instance, Pruning techniques do not compare their results with KD or any other compression technique. This is specifically challenging for researchers and practitioners to decide which compression model technique to use. This section compiles Pruning, KD, and Quantization results for the most commonly used classification datasets: CIFAR 10, CIFAR 100, and ImageNet. We do not present LRMF here since their results are minimal or they use elementary network structures. Pruning and KD results are compiled together in Table~\ref{tab:1} (CIFAR 10), Table~\ref{tab:2} (CIFAR 100), and Table~\ref{tab:3} (ImageNet). Most Pruning techniques concentrate their analysis on the CIFAR 10 and ImageNet datasets, whereas most KD techniques use the CIFAR 100 and ImageNet datasets. 

When analyzing the CIFAR datasets, it becomes evident that some Pruning and KD techniques present a gain in accuracy compared to the baseline model. This phenomenon is common in the context of small datasets, as evidenced in the CIFAR results. However, it is important to note that this phenomenon is less pronounced in the larger ImageNet dataset, with only one case out of the twenty-one reported showing an accuracy gain (Table~\ref{tab:3}). 

In contrast to the typical approach in KD papers, we have chosen not to report the accuracy gain of the compressed model when distilled. Instead, we focus on the loss of the compressed model compared to the baseline model, as well as the reduction in FLOPs and parameters achieved by the compressed model. This decision was made to facilitate a direct comparison between KD and pruning techniques. It is worth noting that we have not included self-distillation results in our analysis, as this process does not involve any reductions in parameters, FLOPs, or execution time.

We also performed experiments using the unchanged ResNet50, ResNet34, ResNet18, and the GReg-2~\cite{wang2020neural} ResNet50 pruned models to analyze how the FLOPs, execution time, and parameters count would be affected in a pruned model when compared to the standard models (Table~\ref{tab:4}). The ResNet34 and ResNet18 models were also included here to evaluate how pruned and distilled networks would perform in one equal setting configuration. We included CPU (i7-7000K) and GPU (RTX 3900) execution times to illustrate the variance in the performance of different devices.

\begin{table*}[!htb]
\caption{Pruning and KD results for the CIFAR 100 dataset. The reported Baseline Model is the larger model used in the compression process. Since Pruning removes weights from the main network and does not involve different architectures, the baseline and compressed models share the same architecture. \dag This result was not available in the paper and was calculated to favor comparison. The table rows are sorted by papers’ published years.}
\setlength{\tabcolsep}{6pt}
\centering
\label{tab:2}

\resizebox{\textwidth}{!}{%
\begin{tabular}{ c | c | c | c | c | c | c | c}


\hline 

& \multirow{2}{4em} {\centering Method}  &  \multirow{2}{4em}{\centering Baseline Model} & \multirow{2}{6em}{\centering Compressed Model}  & \multirow{2}{6em}{\centering Params. Reduction (\%)} & \multirow{2}{6em}{\centering Baseline Model Acc.} & \multirow{2}{6em}{\centering Compressed Model Acc.} & \multirow{2}{4em} {\centering Acc. Drop (\%)} \\
& & & & & & \\
\hline 
 
\multirow{5}{*}{\rotatebox{90}{Pruning}} 

& Liu~\etal~\cite{liu2018rethinking} & DenseNet-BC-100 & DenseNet-BC-100 & 50 & 77.59 $\pm 0.19$ & 76.6 $\pm 0.36$ & 0.99
 \\

& Liu~\etal~\cite{liu2018rethinking} & VGG-19  & VGG-19 & 30 & 71.70 $\pm 0.31$ &  71.96 $\pm 0.36$ & -0.26
  \\

& Liu~\etal~\cite{liu2018rethinking} & VGG-19  & VGG-19 & 50 & 71.70 $\pm 0.31$ &  71.85 $\pm 0.3$ & -0.15
 \\

& GReg-2~\cite{wang2020neural} & VGG-19  & VGG-19 & 88 & 74.02 &  67.75 & 6.47 \\

& DepGraph~\cite{fang2023depgraph} & VGG-19  & VGG-19 & 89 & 73.50 &  70.39 & 3.11 \\


\hline
\hline 

\multirow{9}{*}{\rotatebox{90}{Knowledge Distillation}} & 

Furlanello~\etal~\cite{furlanello2018born}&  Wide-ResNet-28-10 & DenseNet-90-60 & 55~\dag & 80.92 &  83.21 & -2.29 \\ &

Zhang~\etal~\cite{zhang2019your} &  ResNet152 & ResNet18
& 81~\dag & 79.21 &  78.64 & 0.57 \\ &

Zhang~\etal~\cite{zhang2019your} &  ResNet152 & ResNet50
& 58~\dag & 79.21 &  80.56 & -1.35 \\ &

Lin~\etal~\cite{lin2022knowledge} &  ResNet56 & ResNet20
 & 68~\dag & 72.34 &  71.59  & 0.75 \\ &

Yang~\etal~\cite{yang2023knowledge} &  ResNet50 & MobileNetV2
 & 86~\dag & 79.34 &  70.67 & 8.67 \\ &

Li~\etal~\cite{li2023curriculum} &  ResNet50 & MobileNetV2
 & 86~\dag & 79.34 &  68.47  & 10.87 \\ &

Li~\etal~\cite{li2023curriculum} &  ResNet56 & ResNet20 & 68~\dag & 72.34 &  71.19 & 1.15 \\ &

Xu~\etal~\cite{xu2023teacher} &  ResNet152 & ResNet18
 & 81~\dag & 80.91 &  81.27  & -0.36 \\ &

Xu~\etal~\cite{xu2023teacher} &  ResNet152 & ResNet50
 & 58~\dag & 80.91 &  83.09  & -2.18 \\ 

\end{tabular}
}
\end{table*}

\begin{table*}[!htb]
\caption{Pruning and KD results for the ImageNet dataset. The reported Baseline Model is the larger model used in the compression process. Since Pruning removes weights from the main network and does not involve different architectures, the baseline and compressed models share the same architecture. Results that are not available in the papers and could not be estimated are indicated by an em dash (---). \dag result not available in the paper and calculated to favor comparison. The table rows are sorted by papers’ published years.}
\setlength{\tabcolsep}{2pt}
\centering
\label{tab:3}

\resizebox{\textwidth}{!}{%
\begin{tabular}{ c | c | c | c | c | c | c | c | c | c | c | c }


\hline

& \multirow{2}{4em} {\centering Method}  &  \multirow{2}{4em}{\centering Baseline Model} & \multirow{2}{6em}{\centering Compressed Model}  & \multirow{2}{6em}{\centering Params. Reduction (\%)} & \multirow{2}{5em}{\centering Baseline Top1 Acc.} & \multirow{2}{5em}{\centering Compressed Top1  Acc.} & \multirow{2}{6em}{\centering Baseline Model Top5 Acc.} & \multirow{2}{5em}{\centering Compressed Top5  Acc.} &  \multirow{2}{4em}{\centering FLOPs Drop (\%)} & \multirow{2}{5em} {\centering Top1 Acc. Drop (\%)} & \multirow{2}{5em} {\centering Top5 Acc. Drop (\%)} \\
& & & & & & & & & & \\
\hline 
 
\multirow{11}{*}{\rotatebox{90}{Pruning}} 
&  
Liu~\etal~\cite{liu2018rethinking} & VGG-16 & VGG-16 & 30 & 73.37 & 73.68 & --- & --- & --- & -0.31 & ---
  \\

&  
Liu~\etal~\cite{liu2018rethinking} & VGG-16 & VGG-16 & 60 & 73.37 & 73.63 & --- & --- & --- & -0.26 & ---
  \\

&  
Liu~\etal~\cite{liu2018rethinking} & ResNet50 & ResNet50 & 30 & 76.15 & 76.06 & --- & --- & --- & 0.09 & ---
  \\

&  
Liu~\etal~\cite{liu2018rethinking} & ResNet50 & ResNet50 & 60 & 76.15 & 76.09 & --- & --- & --- & 0.06 & ---
  \\


& NISP~\cite{yu2018nisp} & ResNet34  & ResNet34 & 43.68 & --- &  --- & --- &  --- & 43.76 & 0.92 & ---
\\


& NISP~\cite{yu2018nisp} & ResNet50  & ResNet50 & 43.82 & --- &  --- & --- &  --- & 44.01 & 0.89 & ---
\\

& GReg-2~\cite{wang2020neural} & ResNet34  & ResNet34 & --- & 73.31 &  73.61 & --- &  --- & 24.24 & -0.30 & ---
\\

& GReg-2~\cite{wang2020neural} & ResNet50  & ResNet50 & --- & 76.13 &  75.36 & --- &  --- & 56.71 & 0.77 & --- \\

& SRR-GR~\cite{wang2021convolutional} & ResNet50  & ResNet50 & --- & 76.13 &  75.76 & 92.68 &  92.67 & 44.10 & 0.37 & 0.01
\\

& DepGraph~\cite{fang2023depgraph} & ResNet50  & ResNet50 & --- & 76.15 &  76.13 & --- &  --- & 51.82 & 0.77 & --- \\

& DepGraph~\cite{fang2023depgraph} & ViT-B/16  & ViT-B/16 & --- & 81.07 &  79.17 & --- &  --- & 40.91 & 1.9 & --- \\


\hline
\hline 

\multirow{9}{*}{\rotatebox{90}{Knowledge Distillation}} & 

Lin~\etal~\cite{lin2022knowledge}&  ResNet34 & ResNet18 & 46.37~\dag & 73.31 & 72.41 & --- & --- & 50.55~\dag & 0.9 & --- \\ &

Yang~\etal~\cite{yang2023knowledge}&  ResNet34 & ResNet18 & 46.37~\dag & 73.62 & 71.96 & --- & --- & 50.55~\dag & 1.66 & --- \\ &

Yang~\etal~\cite{yang2023knowledge}&  ResNet50 & MobileNetV1 & 83.44~\dag & 76.55 & 72.58 & --- & --- & 86.06~\dag & 3.97 & --- \\ &

Li~\etal~\cite{li2023curriculum}&  ResNet34 & ResNet18 & 46.37~\dag & 73.96 & 71.51 & 91.58 & 90.47 & 50.55~\dag & 2.45 & 1.11 \\ &

DisWOT~\cite{dong2023diswot}&  ResNet34 & ResNet18 & 46.37~\dag & 73.40 & 72.08 & 91.42 & 90.38 & 50.55~\dag & 1.32 & 1.04 \\ &

Li~\etal~\cite{li2023curriculum}&  ResNet34 & ResNet18 & 46.37~\dag & 73.96 & 71.51 & 91.58 & 90.47 & 50.55~\dag & 2.45 & 1.11 \\ &

Huang~\etal~\cite{huang2024knowledge}&  ResNet34 & ResNet18 & 46.37~\dag & 73.31 & 72.49 &  91.42 & 90.71 & 50.55~\dag & 0.82 & 0.71 \\ &

Huang~\etal~\cite{huang2024knowledge}&  ResNet50 & ResNet34 & 14.71~\dag & 80.1 & 78.1 &  --- & --- & 10.51~\dag & 2.0 & --- \\ &

Huang~\etal~\cite{huang2024knowledge}&  ResNet50 & MobileNetV1 & 83.44~\dag & 76.16 & 73.78 & 92.86 & 91.48 & 86.06~\dag & 2.38 & 1.38 \\

\end{tabular}
}
\end{table*}

\begin{table}[!htb]
\caption{Pruning approach comparison with regular ResNet50, ResNet34, and ResNet18 models. The FLOPs Drop column represents the reduction of FLOPs with respect to the regular ResNet50 model. \dag Results generated for comparison.}
\centering
\small
\label{tab:4}

\resizebox{\columnwidth}{!}{%
\begin{tabular}{ c | c | c | c | c }

\hline 

\multirow{2}{*}{Model} &  \multirow{2}{1.2cm}{\centering FLOPs Drop (\%)} & \multirow{2}{1.7cm}{\centering CPU Exec. Time (ms)} & \multirow{2}{1.5cm}{\centering GPU Exec. Time (ms)}  & \multirow{2}{1cm}{\centering Params. Count}  \\
& & & & \\

\hline

GReg-2 R50~\cite{wang2020neural} &  32.89 & 36.16$\pm 3.91$~\dag & 5.67$\pm 0.29$~\dag & 19.63M~\dag \\ 

GReg-2 R50~\cite{wang2020neural} &  67.32 & 29.09$\pm 4.43$~\dag & 5.44$\pm 0.23$~\dag & 11.08M~\dag \\ 

\hline

ResNet50 &  0 & 45.40$\pm 6.25$~\dag & 6.42$\pm 0.28$~\dag & 25.56M~\dag \\ 

ResNet34 &  10.51 & 29.44$\pm 4.09$~\dag & 5.08$\pm 0.70$~\dag & 21.80M~\dag \\ 

ResNet18 &  55.75 & 16.55$\pm 2.54$~\dag & 2.57$\pm 0.20$~\dag & 11.69M~\dag \\ 

\end{tabular}
}
\end{table}

\begin{table*}[!htb]
\centering
\caption{Quantization results for the ImageNet dataset. The reported Baseline Model is the architecture used for model quantization. The compression type is expressed as bits for weights and activations, i.e., 4w8a means the quantization was performed using 4 bits for weights and 8 bits for activations. Results that are not available in the papers are indicated by an em dash (---). The table rows are sorted by papers’ published years.}
\setlength{\tabcolsep}{2pt}
\resizebox{\textwidth}{!}{%
\begin{tabular}{ c | c | c | c | c | c | c | c | c | c }
\hline
\multirow{2}{4em}{\centering Method} & \multirow{2}{4em}{\centering Baseline Model} & \multirow{2}{4em}{\centering Quant. type} & \multirow{2}{6em}{\centering Compression type} & \multirow{2}{6em}{\centering Baseline Top1 Acc.} & \multirow{2}{6em}{\centering Baseline Top5 Acc.} & \multirow{2}{6em}{\centering Compressed Top1 Acc.} & \multirow{2}{6em}{\centering Compressed Top5 Acc.} & \multirow{2}{5em}{\centering Top1 Acc. Drop. (\%)} & \multirow{2}{5em}{\centering Top5 Acc. Drop. (\%)} \\ 
& & & & & & & & & \\ \hline

Mellempudi~\etal~\cite{mellempudi2017ternary} & AlexNet & QAT & 2w8a & 56.83 & --- & 49.04 & --- & 7.79 & --- \\
Mellempudi~\etal~\cite{mellempudi2017ternary} & ResNet50 & QAT & 2w8a & 75.05 & --- & 70.76 & --- & 4.29 & --- \\
Mellempudi~\etal~\cite{mellempudi2017ternary} & ResNet101 & QAT & 2w8a & 77.5 & --- & 73.85 & --- & 3.65 & --- \\ 
Hubara~\etal~\cite{hubara2018quantized} & AlexNet & QAT & 1w2a & 56.6 & 80.2 & 51.03 & 73.67 & 5.57 & --- \\ 
Jacob~\etal~\cite{jacob2018quantization} & ResNet50 & QAT & 8w8a & 76.40 & --- & 74.90 & --- & 1.50 & --- \\ 
Jacob~\etal~\cite{jacob2018quantization} & ResNet150 & QAT & 8w8a & 78.80 & --- & 76.70 & --- & 2.10 & --- \\ 
Jacob~\etal~\cite{jacob2018quantization} & InceptionV3 (RELU6) & QAT & 8w8a & 78.4 ± 0.1 & --- & 75.4 ± 0.1 & --- & 3.00 & --- \\ 
Choi~\etal~\cite{choi2018pact} & AlexNet & QAT & 1w2a & 55.1 & --- & 55.7 & --- & -0.60 & --- \\ 
Choi~\etal~\cite{choi2018pact} & ResNet18 & QAT & 1w2a & 70.2 & --- & 69.8 & --- & 0.40 & --- \\ 
Choi~\etal~\cite{choi2018pact} & ResNet50 & QAT & 1w2a & 76.9 & --- & 76.7 & --- & 0.20 & --- \\ 
Wang~\etal~\cite{wang2019haq} & MobileNetV2 & QAT & 2-6w/a & 71.87 & 90.32 & 71.47 & 90.23 & 0.40 & 0.09 \\ 
Wang~\etal~\cite{wang2019haq} & ResNet50 & QAT & 2-6w/a & 76.15 & 92.86 & 76.14 & 92.89 & 0.01 & -0.03 \\ 
Nagel~\etal~\cite{nagel2019data} & MobileNetV2 & QAT & 8w8a & 71.72 & --- & 70.65 & --- & 1.07 & --- \\ 
Banner~\etal~\cite{banner2019post} & InceptionV3 & PTQ & 4w4a & 77.2 & --- & 68.2 & --- & 9.00 & --- \\ 
Banner~\etal~\cite{banner2019post} & ResNet50 & PTQ & 4w4a & 76.1 & --- & 75.3 & --- & 0.80 & --- \\ 
Banner~\etal~\cite{banner2019post} & ResNet101 & PTQ & 4w4a & 77.3 & --- & 76.9 & --- & 0.40 & --- \\ 
Cai~\etal~\cite{cai2020zeroq} & ResNet50 & PTQ & 8w8a & 77.72 & --- & 77.67 & --- & 0.05 & --- \\ 
Cai~\etal~\cite{cai2020zeroq} & MobileNetV2 & PTQ & 8w8a & 73.03 & --- & 72.91 & --- & 0.12 & --- \\ 
Cai~\etal~\cite{cai2020zeroq} & ShuffleNet & PTQ & 8w8a & 65.07 & --- & 64.94 & --- & 0.13 & --- \\ 
\end{tabular}
}
\label{tab:quantization_comparison}
\end{table*}

Table~\ref{tab:quantization_comparison} presents quantization results for the ImageNet dataset, highlighting the performance impact of several quantization methods on different neural network architectures.
The table focuses on key metrics, including the Top-1 and Top-5 accuracy changes after quantization, providing a clear picture of how quantization affects model performance. Quantization techniques have undergone significant advancements over the years, reflecting the evolution of methodologies and hardware capabilities.
Early quantization methods often involved aggressive quantization, such as those employing binary or 2-bit weights, which leads to substantial accuracy drops \cite{mellempudi2017ternary}.
However, more sophisticated methods using QAT have been developed, exploring mixed precision \cite{choi2018pact, wang2019haq} or flexible precision \cite{wang2019haq} approaches.


\section{Discussion}
\label{disc}

This section discusses the results presented in the previous section, focussing on choosing the best compression technique for a device and how choosing an embedded device configuration can influence the performance of different compression techniques. Additionally, we evaluate the usage evolution of model compression techniques in the literature and discuss trends for each subarea, highlighting recent proposals using one or a combination of different approaches to achieve model compression.

\paragraph{\textbf{Choosing the best technique}} Demonstrating time improvements in model compression models is especially challenging when comparing different approaches because, to the best of our knowledge, no metric studied so far guarantees that one model will be faster than another on all devices. Inference time comparison is straightforward but depends on the evaluated system (e.g., different CPUs, embedded devices, GPUs, or TPUs), which can have specific advantages for particular neural network layers and architectures. Most papers compare the number of parameters and FLOPs, which also do not guarantee a reliable estimation of execution time~\cite{bouzidi2020performance}. The number of parameters can only inform about the memory required to store the 
model in RAM and disk. The computational time of a single layer is composed of the FLOPs count divided by the computation speed, summed with the read and write IO times of the memory involved in the layer's computation~\cite{qi2022paleo}. In addition, the computation time for the whole network is not the summation of the calculation of each individual layer because some architectures, such as the SqueezeNet~\cite{iandola2016squeezenet} and the Inception~\cite{szegedy2015going}, have parallelizable structures.

\begin{figure}[!htb]
\centering
\includegraphics[width=1\linewidth]{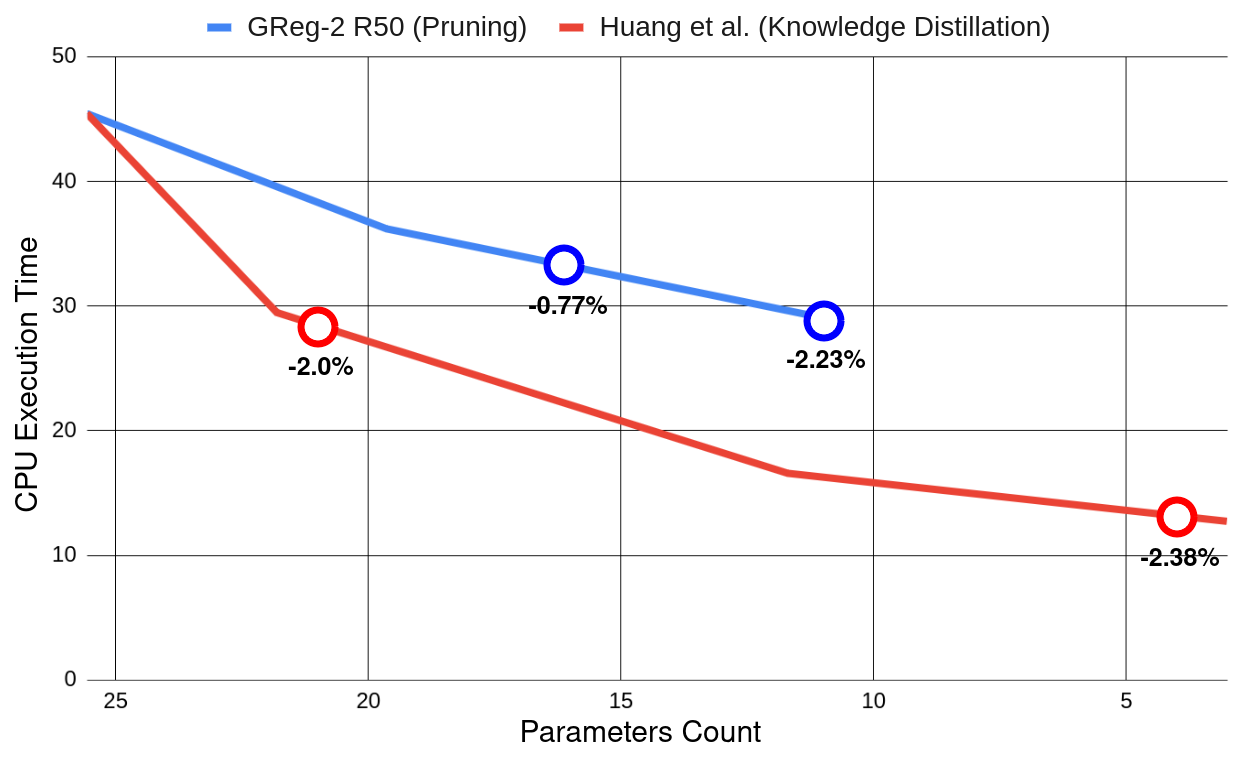}
\vspace*{-5mm}
\caption{Exec. Time vs. Params. Count for Pruning and KD. Points indicate the reduction in Top-1 accuracy compared to the baseline (ResNet50).}
\label{fig:execparams}
\end{figure}

The analysis of Table~\ref{tab:4} corroborates the previous explanation. The GReg-2 pruned versions on a ResNet50 provide a FLOPs reduction of 32.89\% (less pruned model) and 67.32\% (most pruned model) compared to the regular ResNet50. This difference is 23\% and 57\%, respectively, in the parameter size comparison. Despite that, this reduction is not maintained when analyzing the execution times, especially in the GPU. For CPU, the less pruned model only diminishes the execution time by 20\%, and the most pruned model reduces it by 36\%. When compared in a GPU, both reductions drop to 12\% and 15\%, respectively. When analyzing the ResNet34 and ResNet18, it is evident that the execution time reduction is much more significant than the ones observed in the pruned GReg-2 models. The ResNet34 presents 35\% and 21\% reductions in CPU and GPU execution times, while the ResNet18 presents 35\% and 60\% decreases, respectively. Therefore, both models present a higher execution time reduction when compared to equivalent FLOPs-reduced pruned models in our experiments. This effect shows that removing elements in the network but not reducing IO time and not privileging parallelism can lead to a low impact on the execution time. Also, as evidenced by the differences in CPU and GPU time reductions, the gains achieved by smaller or pruned models are highly dependent on the tested device. Figure~\ref{fig:execparams} shows the execution time vs parameters count curves for recent pruning and KD methods in CPU, highlighting the reduction in Top-1 accuracy achieved by each model. The KD method has a more desirable curve, as reductions in parameter count have more impact on execution time. However, small parameter reductions already decrease performance metrics by 2.0\%, whereas larger reductions in GReg-2 only result in a 0.77\% drop in accuracy. On the other hand, the GReg-2 points in the graph indicate that there is a significant impact on metric performance when achieving small decreases in execution time, especially when compared to KD.

In contrast to KD and Pruning techniques, Quantization does not reduce the number of parameters of the compressed network but reduces disk usage, memory allocation, and execution time. RAM and disk usage is reduced by converting 32—or 16-bit float numbers to 8-, 4-, 2-, or 1-bit lower-precision approximations. This also reduces the execution time since operations such as matrix multiplications are improved using integer arithmetic. Usually, the authors discuss execution time or system resource reductions, but these metrics may vary significantly with hardware and implementation. 
Some papers mention and evaluate these aspects, but the variability of evaluated hardware makes direct comparisons challenging \cite{gupta2015deep, baller2021deepedgebench, novac2021quantization}. 
Despite the advances in quantization methods, the classical approaches \cite{jacob2018quantization} are still widely used in research or commercial applications \cite{baller2021deepedgebench, novac2021quantization}, due its easy usability on popular frameworks for embedded systems, such as TensorFlow Lite.

In conclusion, analyzing parameter and FLOPs reductions alone is insufficient to estimate the reduction in execution time, which depends on the compressed architecture and running device. Despite that, both can hint at which model will likely reduce the execution time in different devices. For instance, it is expected that in most devices, the performance of the DepGraph ResNet50 with a 51.82\% drop in FLOPs will be slower than the compressed MobileNetV1 from Yang~\etal~\cite{yang2023knowledge} with 86.06\% FLOPs reduction, but it is challenging to guarantee that the GReg-2 ResNet50 with a 56.71\% drop in FLOPs will be slightly faster than the DepGraph's one in all devices. While execution time analysis is a valuable tool (e.g., in evaluating quantization methods), it is an insufficient metric for guaranteeing the optimal performance of a technique across different devices. Therefore, definitive comparisons between KD, Pruning, and Quantization can only be established on the target device since specific hardware instructions and modules may be highly beneficial for particular situations, such as integer operations and network structures.

\paragraph{\textbf{Compression to Embedded Systems}} One core motivation through compression techniques is using embedded or resource-limited devices. Most techniques presented in this paper allow gains for distinct embedded devices regarding disk usage, memory allocation, and/or execution time.
However, selecting the best methods or combination of techniques may be a complex challenge for these devices.
Embedded devices have different architectures and peculiarities that affect the method's performance, especially when AI accelerators are available.
Indeed, AI accelerator chips made it possible to use complex deep learning architectures on small devices presenting low energy consumption and execution time \cite{jain2020efficient, baller2021deepedgebench, lin2021low, novac2021quantization, bruschi2020enabling}.
When focusing on a specific platform, it is crucial to consider its unique characteristics, such as the processing capabilities of its CPU, GPU, or dedicated AI accelerators, as well as its memory bandwidth and power constraints.

For example, the choice of quantization strategy can significantly influence the model's efficiency on a given device. Some devices handle lower-bit quantization more effectively, whereas others might benefit from mixed-precision approaches. Additionally, hardware-specific optimizations, such as using tensor cores on NVIDIA GPUs or the Edge TPU's capabilities on Google devices, can further enhance performance.
Several recent research papers have investigated the performance trade-offs between different hardware platforms, including AI accelerators~\cite{jain2020efficient, baller2021deepedgebench, lin2021low}, microcontrollers~\cite{novac2021quantization, bruschi2020enabling}, and FPGAs~\cite{chen2020learning, ducasse2021benchmarking}.

As previously discussed, certain hardware platforms can still be optimized for specific network architectures, impacting the performance comparison of other compression techniques on different devices. For example, some devices might better handle the sparse matrices produced by pruning or the simplified architectures generated through knowledge distillation.

\begin{figure}[!htb]
\centering
\includegraphics[width=1\linewidth]{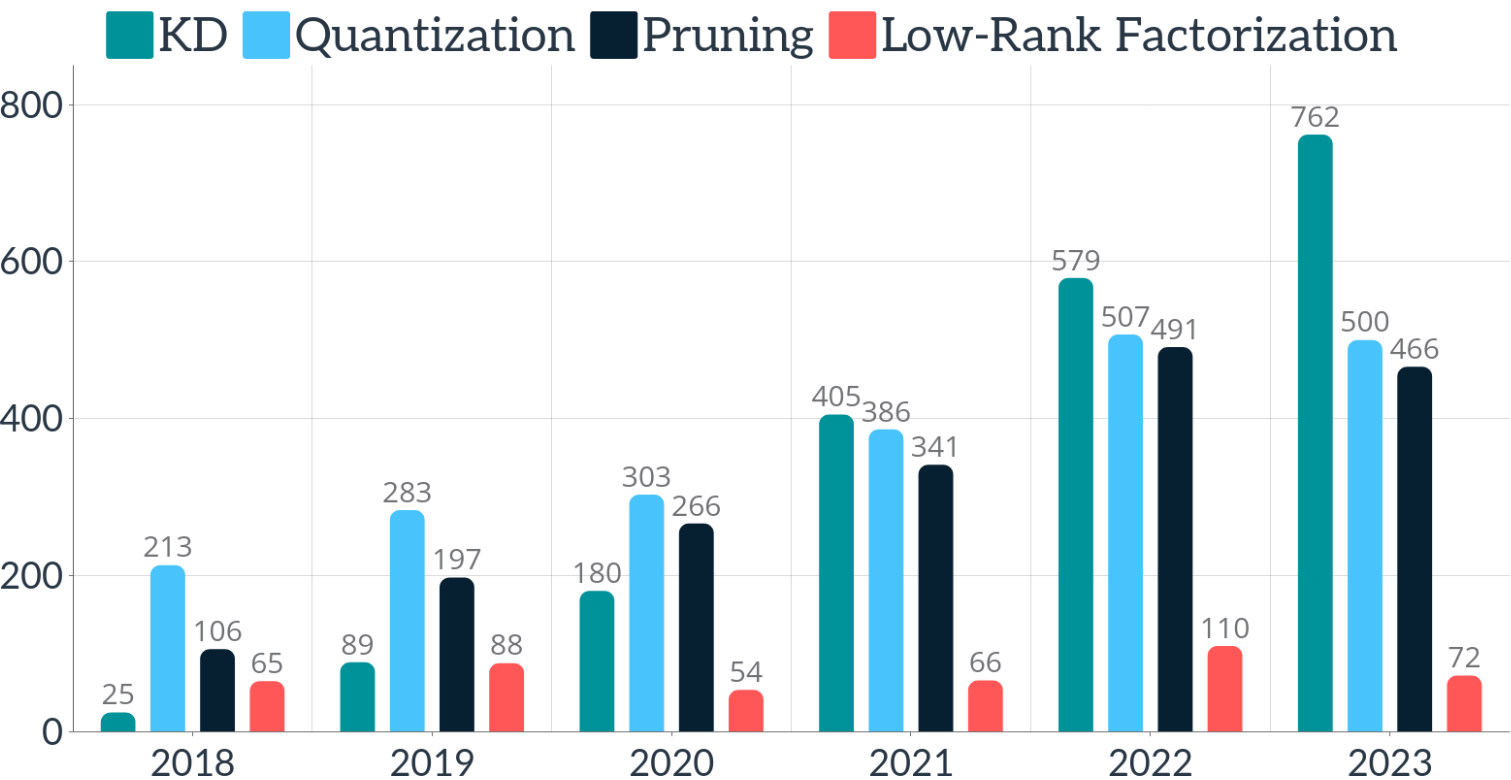}
\vspace*{-5mm}
\caption{Evolution of Usage of Model Compression by Year. We report the number of found papers published using or proposing model compression techniques by subarea by year. The search was conducted in Scopus and included years from 2018 to 2023.}
\label{fig:modcomp}
\end{figure}

\paragraph{\textbf{Prominent subareas}} We monitor the evolution of interest in each area for Computer Vision, as seen in Figure~\ref{fig:modcomp} using the searching phrases discussed in Section~\ref{sec:met}. Analyzing the total number of papers and growth curve for each compression subarea, we observe that Knowledge Distillation is the most utilized technique for model compression in Computer Vision and exhibits the fastest growth compared to other methods. The number of papers incorporating Quantization and Pruning was nearly identical in 2022 and 2023, with a 1.3\% and 5.1\% decrease for each, respectively. In contrast, the number of papers using KD increased by 31.6\% in 2023, totaling 762 papers. This increase can be attributed to the favorable metric results and execution time reduction achieved through distinct hardware using KD and its capability to enhance the generalization of the models. Improving generalization is also beneficial for scenarios not involving compression~\cite{zhang2019your, yun2020regularizing, yang2023knowledge}. These papers also help to boost the recent interest in KD, although this usage do not provide compression results. Low-Rank Matrix Factorization maintained a low usage compared to other techniques, especially after 2019, and experienced a reduction from 2022 to 2023 (34.54\%). These papers rarely relate to compression and have been used primarily in other computer vision applications, such as improving visual data representation~\cite{wang2023self, liu2020research}.

\paragraph{\textbf{Trends}} The compression algorithms are not mutually exclusive and can be combined to reduce parameter count and processing time. For example, recent techniques combine Knowledge Distillation with Quantization~\cite{ji2022neural, huang2022compressing} or Pruning with Quantization~\cite{liberatori2022yolo}. This combination rarely involves Low-Rank Matrix Factorization due to the presented limitations on CNNs. The high usage of Transformer architectures in the Natural Language Processing area favored the appearance of recent low-rank factorization techniques~\cite{hu2021lora,hsu2022language}, which tends to also happen for Computer Vision applications.

For Knowledge Distillation, the use of self-distillation to improve the performance of modern architectures is a recent trend~\cite{kang2023distilling, yun2020regularizing, yang2023knowledge}, although not having the goal of resulting in a compressed model. Also, KD has been used to improve visual representation in unsupervised settings for images~\cite{chen2023sssd, song2023multi} and videos~\cite{wang2023masked}. Novel techniques that involve compression try to adapt KD to new architectures, such as for transformer-based detection~\cite{chang2023detrdistill} and classification models~\cite{chen2022dearkd}. Additionally, Huang~\etal~\cite{huang2024knowledge} proposed a diffusion model to denoise student features before distilling knowledge from the teacher.    

For pruning, novel prominent techniques are based on structural pruning due to its ability to group and remove elements from the network, providing better results in terms of compression~\cite{lin2022knowledge, fang2023depgraph, huang2023cp3}. Adapting pruning for architectures not tested in the paper can be very difficult, and fully automatic methods such as DepGraph~\cite{fang2023depgraph} can facilitate the process. Additionally, pruning has been recently applied for other tasks, such as on 3D point-based neural networks~\cite{huang2023cp3} and diffusion models~\cite{fang2024structural}.

With modern quantization methods achieving minimal errors, the actual trends concentrate on Vision Transformers (ViT), and recent researches have been exploring this opportunity~\cite{liu2021post, li2022q, yuan2022ptq4vit, du2024model}.
It is an interesting research option with several challenges since the difficulties faced by binary quantization in Vision Transformers include the significant reduction in parameter precision, leading to a drop in model performance. Additionally, weight oscillation during training is problematic, as is the adaptation of low-precision representations for ViTs, resulting in precision loss. When quantizing pre-trained models to extremely low representations, the non-convex loss landscape is another challenge, along with the need for specific strategies for the complex task of binary quantization in ViTs. The unbalanced distribution of activation values after non-linear functions, such as Softmax and GELU, presents challenges in quantization, as does the inter-channel variability in quantizing outputs of normalization layers.

\section{Case Study}
\label{caseStudies}

We prepared a case study example using different datasets and backbones, in which we apply isolated and combined compression techniques to analyze the accuracy, parameter count, and processing time of various scenarios for these datasets. The case study is available in our GitHub\footnote{The URL will be released after the acceptance of this paper.}.

\section{Conclusion}
\label{secConclusion}

Recent Computer Vision methods rely on large deep learning networks, which are generally unsuitable for devices with limited computer power, such as embedded systems. Model compression techniques aim to balance the application's performance with model size and inference time, allowing powerful models to be deployed in scenarios with limited resources.

This survey investigates various subareas of model compression techniques for computer vision applications. We explored subcategories of algorithms, highlighting the distinctions within each proposed method and providing pseudocode to facilitate a deeper understanding. We reported and discussed trends, conducted a performance comparison of compression techniques, and performed experiments to evaluate FLOPs, parameter count, and execution time in different hardware. Our analysis suggests that FLOPs, execution time, and parameter count can serve as initial indicators for choosing the optimal model for a specific device. However, definitive conclusions of the best-performing technique require benchmarking directly on the target device. This happens because particular hardware characteristics, such as instruction sets and modules, can significantly influence the performance of specific network architectures and compressed models.

Although compression techniques have found extensive application in NLP Transformer-based architectures, their exploration in computer vision transformer-based models is just beginning, presenting a significant gap compared to their utilization in NLP. Addressing this gap could unlock the potential for leveraging underutilized techniques, such as Low-Rank Matrix Factorization, to achieve further gains.

\bibliographystyle{cag-num-names}
\bibliography{biblio}



\end{document}